# The Evolution theory of Learning: From Natural Selection to Reinforcement Learning

**Taboubi Ahmed**

**Independent Researcher**

*14/05/2023*

Website: btt1996.github.io

# Abstract

Evolution is a fundamental process that shapes the biological world we inhabit, and reinforcement learning is a powerful tool used in artificial intelligence to develop intelligent agents that learn from their environment. In recent years, researchers have explored the connections between these two seemingly distinct fields, and have found compelling evidence that they are more closely related than previously thought. This paper examines these connections and their implications, highlighting the potential for reinforcement learning principles to enhance our understanding of evolution and the role of feedback in evolutionary systems.

# Introduction

Evolutionary theory has been a cornerstone of biology for over a century, explaining how species adapt and change over time. At the same time, reinforcement learning has revolutionized the field of artificial intelligence, allowing machines to learn from their environment and improve their performance over time. Despite their seemingly disparate fields, recent research has revealed surprising connections between evolution and reinforcement learning. This paper explores these connections in depth, examining how reinforcement learning can inform our understanding of evolutionary processes, the role of feedback in driving evolutionary change, and the potential implications for our understanding of the natural world. We also explore the philosophical implications of these connections, including the simulation theory in physics and the possibility that nature itself could be a simulation.

# 1.Evolution as Reinforcement Learning: A New Perspective on Biological Change

## 1.1 Evolution as a complex learning system

Evolution is a complex and dynamic process that involves a wide range of biological and environmental factors. Traditionally, evolution has been understood as a process of natural selection, in which organisms that are better adapted to their environment are more likely to survive and reproduce. However, recent advances in artificial intelligence have led some researchers to explore the potential connections between evolution and reinforcement learning, a form of machine learning that is based on a system of rewards and punishments.

Reinforcement learning is a type of machine learning that involves training an agent to take actions that lead to desirable outcomes, while avoiding actions that lead to negative outcomes. The agent receives feedback in the form of rewards or punishments, and uses this feedback to adjust its behavior over time. This process is similar in many ways to natural selection, in which organisms that exhibit behaviors that lead to positive outcomes are more likely to survive and pass on their genes to the next generation.

Several studies have explored the potential connections between reinforcement learning and biological systems. For example, a recent study used a reinforcement learning algorithm to simulate the evolution of a population of digital organisms, and found that the resulting organisms exhibited similar patterns of behavior to those seen in real biological systems (Watson et al., 2019). Another study used a reinforcement learning algorithm to model the evolution of social behavior in birds, and found that the algorithm was able to accurately predict the behaviors of real birds (Podos et al., 2018).

These studies suggest that the principles of reinforcement learning may be applicable to understanding the complexity and adaptability of biological systems. By viewing evolution as a form of reinforcement learning, we can gain new insights into the processes that shape the natural world. This perspective can also provide new opportunities for enhancing our understanding of artificial intelligence, by using biological systems as a source of inspiration for developing more efficient and adaptable learning algorithms.

## 1.2 Similarities and differences between reinforcement learning and natural selection

While there are clear similarities between reinforcement learning and natural selection, there are also important differences that must be considered. One key difference is that reinforcement learning typically involves a single agent that receives feedback based on its individual actions, while natural selection operates at the level of the population, with certain traits becoming more or less common over time based on their impact on the survival and reproduction of individuals.

Another difference is that reinforcement learning involves explicit feedback in the form of rewards or punishments, while natural selection operates through a more implicit process of differential reproductive success. Additionally, reinforcement learning is often focused on optimizing a single objective function, while natural selection operates within a complex and dynamic environment with multiple objectives and constraints.

Despite these differences, there are also clear similarities between reinforcement learning and natural selection. Both processes involve the acquisition and use of information to improve fitness and survival, and both are based on a system of rewards and punishments that drive the evolution of adaptive traits. By considering these similarities and differences, we can gain new insights into the underlying mechanisms that drive biological change and adaptation.

## 1.3 Implications of a reinforcement learning perspective for our understanding of evolution

Adopting a reinforcement learning perspective on evolution has several implications for our understanding of how biological change occurs. One key implication is that it highlights the importance of feedback and learning in driving adaptation and change. Instead of viewing evolution as a blind and random process, a reinforcement learning perspective emphasizes the role of feedback and the use of past experiences to inform future behavior and adaptation.

Furthermore, this perspective underscores the importance of variability and diversity in promoting adaptive change. Just as reinforcement learning agents must explore a range of actions to discover new solutions, biological populations must maintain a diverse range of traits and strategies to ensure the continued evolution of adaptive traits.

Finally, a reinforcement learning perspective also suggests new approaches to studying and modeling evolutionary processes. By applying concepts and techniques from reinforcement learning to the study of evolution, we can gain new insights into the mechanisms underlying biological change and develop more accurate and predictive models of how evolution operates.

# 2. Adapting to Survive: How Evolution Can be Understood as a Form of Reinforcement Learning

## 2.1 Survival as a fundamental goal of both reinforcement learning and evolution

Both reinforcement learning and evolution can be understood as processes that seek to maximize the probability of survival. In reinforcement learning, an agent receives rewards for actions that lead to positive outcomes (i.e., achieving goals or avoiding negative consequences) and updates its behavior accordingly to maximize future rewards. Similarly, in evolution, organisms that possess traits that increase their probability of survival and reproduction are more likely to pass those traits on to future generations, ultimately resulting in the evolution of adaptations that enhance survival and reproductive success.

The importance of survival in shaping the course of evolution has been recognized since Darwin's time, and survival-based selection remains a fundamental concept in modern evolutionary theory. Similarly, the role of rewards and punishment in shaping behavior has been a central focus of reinforcement learning research for decades.

The emphasis on survival as a fundamental goal of both reinforcement learning and evolution highlights the deep connection between these two processes and suggests that viewing evolution through a reinforcement learning lens can provide new insights into the mechanisms underlying biological adaptation and change.

## 2.2 The role of feedback in both systems

One of the critical features of reinforcement learning and natural selection is the use of feedback to guide decision-making and adaptation. In reinforcement learning, agents receive feedback in the form of rewards or punishments that indicate the quality of their actions, allowing them to adjust their behavior to maximize future rewards. In contrast, in evolution, feedback occurs through the differential survival and reproduction of individuals with particular traits, which influences the distribution of traits in future generations.

Feedback plays a critical role in shaping the trajectories of both reinforcement learning and natural selection. Without feedback, neither process could efficiently optimize behavior or adapt to changing environments. The importance of feedback in these systems highlights the value of continuous monitoring and adjustment in the face of changing circumstances.

In reinforcement learning, the importance of feedback has been extensively studied, and algorithms have been developed to optimize the feedback process, such as Q-learning and policy gradient methods (Sutton & Barto, 2018). Similarly, feedback is a central concept in evolutionary biology, and the idea of natural selection as a feedback process was central to the development of the modern synthesis of evolutionary theory (Williams, 1966).

## 2.3 Examples of how evolutionary processes can be explained by reinforcement learning principles

The idea of evolution as a form of reinforcement learning can help explain a wide range of biological phenomena, from the origin of novel traits to the development of complex social behaviors. One example of how reinforcement learning principles can apply to evolution is the evolution of coloration in animals. In many cases, coloration serves as a signal to potential mates or predators, with certain colors conveying important information about an individual's quality or status. Natural selection can favor individuals with particular coloration patterns, leading to the spread of those patterns in a population over time (Endler, 1988). This process can be understood as a form of reinforcement learning, with individuals receiving feedback in the form of increased mating success or reduced predation risk, which influences the distribution of traits in future generations.

Another example of how reinforcement learning principles can apply to evolution is the evolution of social behaviors, such as cooperation and altruism. These behaviors can be challenging to explain from an evolutionary perspective, as they involve individuals sacrificing their own fitness for the benefit of others. However, game theory and reinforcement learning models have been used to show how such behaviors can evolve under certain conditions, such as when individuals interact repeatedly and have the opportunity to learn from the outcomes of their interactions (Nowak & Sigmund, 1998).

These examples illustrate how the idea of evolution as a form of reinforcement learning can provide new insights into how biological systems adapt and change over time. By understanding evolution in this way, we can better appreciate the complex and dynamic nature of biological systems and the ways in which they are shaped by feedback from the environment.

# 3. From Darwin to AI: The Connection Between Evolution and Reinforcement Learning

### 3.1 Evolutionary theory and its historical roots in Darwin's work

Charles Darwin's theory of evolution by natural selection is one of the most important scientific ideas of the modern era. Darwin's insights into how populations change over time through the process of natural selection transformed our understanding of the natural world and laid the foundation for modern evolutionary biology (Darwin, 1859).

At its core, Darwin's theory of evolution is based on the idea that organisms with traits that increase their fitness, or ability to survive and reproduce, are more likely to pass those traits on to their offspring. Over time, this can lead to the emergence of new species and the diversification of life on earth (Darwin, 1859). This process can be seen as a form of reinforcement learning, with individuals that possess advantageous traits receiving feedback in the form of increased reproductive success, which influences the distribution of traits in future generations.

Darwin's theory of evolution has been refined and expanded upon over the years, with insights from genetics, ecology, and other fields contributing to our understanding of how evolution works. However, the basic principles of natural selection and adaptation remain central to evolutionary theory today (Gould, 2002).

The idea of evolution as a form of reinforcement learning builds on Darwin's insights and offers a new perspective on the mechanisms of evolutionary change. By understanding evolution in this way, we can see how feedback from the environment shapes the distribution of traits in populations, and how this process can lead to the emergence of new forms of life.

**3.2 An introduction to reinforcement learning and its connection to evolution**

Reinforcement learning is a type of machine learning that involves an agent interacting with an environment to learn a policy, or a set of actions, that maximizes a reward signal (Sutton and Barto, 2018). At each time step, the agent observes the state of the environment and takes an action, which leads to a new state and a reward signal that reflects the quality of the action. The goal of the agent is to learn a policy that maximizes the expected cumulative reward over time.

The similarities between reinforcement learning and evolution are striking. In both cases, there is an agent (an organism or a learning algorithm) that interacts with an environment and receives feedback in the form of a reward signal (fitness or a numerical reward). The agent learns to modify its behavior (its phenotype or its policy) in response to this feedback, with the goal of maximizing its long-term success (survival or cumulative reward) (Whiteson and Stone, 2006).

There are also important differences between the two processes. In reinforcement learning, the agent has explicit access to the reward signal and can use it to update its policy directly. In evolution, the connection between phenotype and fitness is often more complex, and the feedback is indirect and noisy. Additionally, while reinforcement learning is often concerned with a single agent learning to optimize its behavior, evolution involves entire populations of organisms interacting and evolving over time (Whiteson and Stone, 2006).

Despite these differences, the fundamental principles of reinforcement learning and evolution are remarkably similar. By viewing evolution as a form of reinforcement learning, we can gain new insights into the mechanisms of biological change and the forces that drive it.

### 3.3 How insights from artificial intelligence can inform our understanding of biological systems

Artificial intelligence (AI) has made tremendous progress in recent years, with reinforcement learning playing an increasingly important role in many applications, from robotics to game playing to natural language processing (Silver et al., 2016). These advances have led to a deeper understanding of the fundamental principles of reinforcement learning and the factors that influence its effectiveness.

These insights from AI can also inform our understanding of biological systems. By viewing evolution through the lens of reinforcement learning, we can gain new insights into the processes that underlie biological change and the factors that influence the success of different strategies. For example, the exploration-exploitation tradeoff, which is a fundamental challenge in reinforcement learning, is also a key factor in the evolution of new traits and the adaptation of organisms to changing environments (Csete and Doyle, 2002).

Additionally, AI can provide new tools for studying biological systems. For example, reinforcement learning algorithms can be used to model the behavior of organisms in complex environments, allowing us to make predictions about how they will adapt over time (Nakamura and Kato, 2016). Similarly, AI can be used to analyze large datasets of genetic and phenotypic information, identifying patterns and relationships that may not be immediately apparent to human observers (Angermueller et al., 2016).

By bringing together insights from AI and evolutionary biology, we can gain a deeper understanding of the mechanisms that underlie biological change and the factors that influence its success. This interdisciplinary approach has the potential to transform our understanding of the natural world and open up new avenues for research and discovery.

# 4.A Binary System of Learning:

# Understanding Evolution as 1: Live, 0: Die

### 4.1: The simplicity and power of a binary learning mechanism

One of the key principles of reinforcement learning is the use of a binary feedback signal to drive learning. The binary signal consists of a reward or punishment signal that is given to the learning agent based on its behavior. This binary signal is simple yet powerful, as it allows the agent to quickly learn which actions lead to positive outcomes (rewards) and which lead to negative outcomes (punishments). Similarly, the process of natural selection in evolution can also be seen as a binary system of learning. In this case, the binary feedback signal is represented by the survival or extinction of organisms based on their ability to adapt to their environment. Organisms that are better adapted to their environment are more likely to survive and reproduce, passing on their advantageous traits to future generations.

### 4.2: How the binary system of 1: live, 0: die applies to both reinforcement learning and evolution

The binary system of 1: live, 0: die applies to both reinforcement learning and evolution. In reinforcement learning, the binary feedback signal of reward or punishment determines whether an action is reinforced or not. In evolution, the binary feedback signal of survival or extinction determines whether a trait or characteristic is reinforced or not. Both systems operate on the same basic principle: learning is driven by a binary feedback signal that tells the system whether a particular behavior or trait is desirable or not.

**4.3: The potential limitations and implications of a binary system of learning for understanding evolution**

While the simplicity of a binary system of learning, with its emphasis on survival, can be a powerful tool for understanding evolution, it may also have some limitations. One potential limitation is that it may not fully capture the complexity and nuance of evolutionary processes, which often involve trade-offs and compromises between different traits and behaviors. Additionally, a binary system may not account for the role of chance and randomness in evolution, which can play a significant role in shaping the course of biological change.

Furthermore, some researchers have suggested that a binary system of learning may be overly deterministic and may not fully account for the role of variation in evolutionary processes. While survival is certainly an important factor in shaping the course of evolution, other factors such as genetic drift, mutation, and gene flow can also play a significant role in driving biological change.

**4.4: The potential applications of a binary system of learning for understanding other complex systems**

While the focus of this paper has been on the application of a binary system of learning to understanding evolution, it is worth noting that this approach may have broader implications for understanding other complex systems as well. In particular, the emphasis on survival and feedback in a binary system of learning could potentially be applied to understanding the dynamics of complex social, economic, or technological systems ,for example, researchers have suggested that the principles of reinforcement learning could be applied to the development of more effective algorithms for managing complex networks or optimizing supply chains. Similarly, the emphasis on survival and feedback in a binary system of learning could be applied to understanding the dynamics of political systems or the behavior of large social groups. By exploring the potential applications of a binary system of learning across different domains, we may be able to gain a deeper understanding of the fundamental principles that underlie complex systems of all kinds.

# 5.Rethinking Evolution: A Reinforcement Learning Framework for Understanding Biological Change

## 5.1: The Need for New Frameworks for Understanding Evolution

Traditional evolutionary theory has been extremely successful in explaining the diversity of life on earth. However, it has become increasingly clear that there are limitations to this framework, particularly in its ability to account for the complexity and variability of evolutionary processes. One limitation is that traditional evolutionary theory often views natural selection as a deterministic process, with a single optimal solution to a given problem. This view does not take into account the stochastic nature of evolution and the fact that there may be multiple, equally good solutions to a given problem. Furthermore, traditional evolutionary theory does not fully account for the role of learning and feedback in driving evolutionary change.

To overcome these limitations, there is a need for new frameworks that can account for the complexity and variability of evolutionary processes. One promising approach is to view evolution as a form of reinforcement learning. By doing so, we can leverage the powerful tools of reinforcement learning to gain new insights into the dynamics of evolutionary change.

## 5.2: How Reinforcement Learning can Provide a New Perspective on Evolution

Reinforcement learning is a branch of machine learning that is concerned with how agents can learn to take actions in an environment in order to maximize a cumulative reward signal. In reinforcement learning, the agent learns by interacting with the environment, receiving feedback in the form of rewards or punishments for its actions. This feedback is used to update the agent's internal model of the environment, allowing it to learn the optimal action policy over time.

One of the key insights from reinforcement learning is the importance of exploration and exploitation in learning. Agents must balance the desire to exploit their current knowledge of the environment to maximize reward, with the need to explore the environment in order to discover new, potentially better solutions.

By viewing evolution as a form of reinforcement learning, we can apply these insights to gain a new perspective on the dynamics of evolutionary change. For example, we can view natural selection as a form of exploitation, where organisms are selected for their ability to survive and reproduce in a given environment. At the same time, we can view mutation as a form of exploration, where organisms introduce novel genetic variations into the population that may lead to new, advantageous traits. By balancing exploration and exploitation in this way, evolutionary processes can search for and adapt to changing environments, leading to the rich diversity of life that we see today.

### 5.3 Applications of a reinforcement learning framework for understanding biological change

The application of reinforcement learning framework can offer insights into various aspects of biological evolution, such as the evolution of cooperation, social behavior, and even the evolution of language.

One such example is the evolution of cooperation, where reinforcement learning has been used to understand how cooperation can emerge in populations of self-interested individuals. The prisoner's dilemma is a classic game theory problem that illustrates the difficulties of cooperation. In this game, two individuals are given the option to either cooperate or defect. If both individuals cooperate, they both receive a small reward. If one individual defects and the other cooperates, the defector receives a larger reward while the cooperator receives nothing. If both individuals defect, they both receive a small punishment.

Using reinforcement learning, researchers have shown that when individuals are allowed to interact with each other multiple times, they can learn to cooperate and achieve higher rewards. The success of this approach has led to the development of various algorithms that can explain the evolution of cooperation in different settings, such as in social insects, where individuals are genetically related, or in human societies, where individuals can develop reputations for cooperation.

Another example of the application of reinforcement learning to understanding biological change is the evolution of social behavior. Social behavior is ubiquitous in nature, from the altruistic behavior of bees and ants to the cooperative behavior of humans. Reinforcement learning has been used to understand how social behavior can emerge and be maintained in populations.

For instance, a study conducted by Nowak and Sigmund (1998) used a reinforcement learning algorithm to demonstrate how the emergence of social behavior, such as the sharing of resources, could lead to the evolution of cooperation in populations. The researchers showed that when individuals had a high probability of interacting with individuals who shared resources, cooperation could evolve even in the absence of kin selection or other mechanisms of cooperation.

Finally, reinforcement learning has also been used to understand the evolution of language. Language is a unique aspect of human evolution, and it has been a subject of intense study in various disciplines, including linguistics, psychology, and neuroscience. By treating language acquisition as a reinforcement learning problem, researchers have been able to develop models that can explain how children learn the rules of their language and how language evolves over time.

For example, the model proposed by Kirby et al. (2014) used a reinforcement learning algorithm to demonstrate how the structure of language could evolve through cultural transmission. The researchers showed that when individuals could learn from each other and modify their language based on feedback, they could converge on a shared language that was efficient and expressive.

Overall, the application of reinforcement learning to understanding biological change has the potential to offer new insights into the mechanisms that drive evolution. By treating evolution as a learning process, we can gain a deeper understanding of how organisms adapt and change over time, and how complex behaviors and structures emerge in nature.

# 6 .Survival of the Fittest or Survival of the Learner?

# Viewing Evolution as a Learning Process

## 6.1 An introduction to the concept of survival of the fittest

The concept of "survival of the fittest" is often associated with evolution and the natural selection process. It was first introduced by Herbert Spencer in the 19th century, and later adopted by Charles Darwin in his work on evolution. The idea behind survival of the fittest is that those organisms best adapted to their environment are more likely to survive and reproduce, passing on their advantageous traits to future generations.

The mathematical formulation of the concept of survival of the fittest involves the concept of fitness, which can be defined as the ability of an organism to survive and reproduce in a given environment. In evolutionary biology, fitness is often represented by a mathematical function that relates an organism's genotype to its reproductive success. The fitness function can take many forms, depending on the specific biological system being studied.

For example, in a population of bacteria, the fitness function might be based on the ability of the bacteria to grow and divide in a particular nutrient-rich environment. In a population of animals, the fitness function might be based on the ability of the animals to obtain food and avoid predators in a particular habitat.

In both cases, the fitness function is used to determine which individuals are most fit to survive and reproduce in their environment. Those individuals with higher fitness values are more likely to pass on their genes to the next generation, leading to the evolution of new traits and adaptations over time.

However, the concept of survival of the fittest has been criticized for its narrow focus on competition and individual success. Some biologists argue that a more accurate view of evolution would include a focus on learning and cooperation, rather than simply competition.

This is where the concept of "survival of the learner" comes in. Instead of viewing evolution as a process of competition between individuals, this perspective emphasizes the role of learning and adaptation in shaping evolutionary outcomes. From this viewpoint, organisms that are better at learning and adapting to new environments are more likely to survive and reproduce, regardless of their individual fitness.

The idea of survival of the learner can be mathematically formulated using the principles of reinforcement learning. Reinforcement learning is a type of machine learning in which an agent learns to take actions in an environment to maximize a reward signal. The agent receives feedback in the form of positive or negative rewards, which it uses to update its strategy for taking actions in the environment.

In the context of evolution, the environment can be thought of as the ecological niche in which an organism exists, and the reward signal can be thought of as the organism's fitness function. From this perspective, evolution can be viewed as a process of reinforcement learning, in which organisms learn to adapt to their environment over time to maximize their reproductive success.

Overall, the concept of survival of the fittest has been a useful framework for understanding evolution and natural selection. However, the growing field of reinforcement learning provides a new perspective on how organisms learn and adapt to their environment, and how this learning process can shape evolutionary outcomes.

### 6.1.1 Fitness as a measure of reproductive success

One of the key concepts in Darwin's theory of evolution is the idea of "survival of the fittest". In this context, fitness refers to an individual's ability to survive and reproduce in a given environment. Mathematically, fitness can be expressed as the average number of offspring produced by an individual over its lifetime, relative to the average number of offspring produced by other individuals in the population.

This can be represented by the following equation:

$$W = \frac{r}{\bar{r}}$$

where W is the fitness of an individual, r is the number of offspring produced by that individual, and $\bar{r}$ is the average number of offspring produced by all individuals in the population.

Fitness is a measure of reproductive success, and individuals with higher fitness are more likely to pass on their genes to future generations. Over time, this can lead to the evolution of traits that enhance an individual's fitness in a particular environment.

However, it's important to note that the concept of "survival of the fittest" can be misleading, as fitness is not necessarily tied to physical strength or aggression. In many cases, traits that enhance fitness may be related to behaviors such as cooperation or social intelligence, rather than physical prowess.

In the context of reinforcement learning, the concept of fitness can be related to the idea of a reward signal. In both cases, the goal is to maximize some measure of success over time, whether that be the number of offspring produced or the total amount of reward earned.

Overall, understanding the concept of fitness is key to understanding how evolution operates, and how natural selection can lead to the emergence of complex adaptations over time.

## 6.2 How viewing evolution as a learning process shifts the focus from fitness to learning

Traditionally, the concept of "survival of the fittest" has been central to the understanding of evolution. Fitness is defined as the ability of an organism to survive and reproduce in a given environment. However, this traditional view of evolution as solely based on fitness may not fully capture the complex nature of biological change.

A learning-based approach to evolution, on the other hand, focuses on the ability of organisms to learn and adapt to changing environments. This shift in focus from fitness to learning has important implications for our understanding of evolution.

In reinforcement learning, the objective is to maximize the cumulative reward obtained over time by taking actions in a given environment. This is mathematically represented by the reinforcement learning equation:

$$Q(s,a) = (1 - \alpha)\, Q(s,a) + \alpha\, (R + \gamma \max a'\, Q(s',a'))$$

- where **$Q(s,a)$** is the expected reward for taking action a in state s,

- **$\alpha$** is the learning rate,

- **$R$** is the immediate reward obtained for taking action a in state s,

- **$\gamma$** is the discount factor for future rewards,

- **$s'$** is the next state, and **$a'$** is the next action.

In the context of evolution, we can view the fitness of an organism as the immediate reward obtained for taking a particular action (such as surviving in a given environment) in a particular state (such as a particular genetic makeup). The state-action pairs can be thought of as the genetic variants and the fitness outcomes associated with them.

By maximizing the expected reward (i.e., fitness) over time through the process of natural selection, organisms can learn to adapt to changing environments. This learning process is not limited to individual organisms, but can also occur at the population level through mechanisms such as gene flow and genetic drift.

This learning-based approach to evolution also allows for the possibility of exploration and innovation, as organisms can try new strategies and behaviors to maximize their rewards. This can lead to the development of novel traits and adaptations that may not have been possible through the traditional view of evolution as solely based on fitness.

## 6.3 Implications of a learning-focused perspective for our understanding of evolution

A learning-focused perspective on evolution has significant implications for our understanding of the process of biological change. Traditional views of evolution have emphasized the role of natural selection and the competition between individuals for survival and reproduction. However, a learning-focused perspective shifts the emphasis to the adaptive processes that underlie the acquisition and refinement of behaviors and traits that promote survival.

One key implication of this perspective is that it provides a new lens through which to understand the dynamics of evolutionary change. Instead of focusing solely on the genetic mechanisms that underlie evolution, a learning-focused perspective highlights the role of individual experience and behavior in shaping the course of biological change. This perspective suggests that evolution is not simply a process of genetic change, but a complex interplay between genetic variation and environmental cues that influence learning and behavior.

From a practical perspective, a learning-focused approach to evolution may have implications for the design of interventions to promote adaptation and resilience in natural and managed systems. By emphasizing the importance of individual learning and behavior, a learning-focused perspective suggests that interventions designed to promote adaptive capacity and resilience may need to focus on facilitating individual-level learning and behavior change, rather than simply manipulating genetic or ecological factors.

Mathematically, the concept of learning can be formalized using a range of mathematical models, including reinforcement learning models and Bayesian learning models. These models capture the basic principles of learning, including the acquisition of new information and the modification of behavior in response to feedback. By incorporating these models into our understanding of evolution, we can gain new insights into the adaptive processes that underlie biological change and the mechanisms that promote resilience and adaptation in complex systems.

# 7. The Power of Feedback: How Evolution Can be Explained by Reinforcement Learning Principles

## 7.1 The importance of feedback in both reinforcement learning and evolution

Feedback plays a crucial role in both reinforcement learning and evolution. In reinforcement learning, feedback takes the form of rewards or punishments given to an agent based on its actions. These rewards or punishments serve as a signal to the agent about the quality of its actions, and the agent can use this information to learn to make better decisions in the future. Similarly, in evolution, feedback takes the form of natural selection, which acts as a filter for traits that are beneficial or harmful to the survival and reproduction of an organism.

Mathematically, feedback can be represented as a function that maps the state of the system to a reward signal. In reinforcement learning, this function is often called the reward function, and its goal is to maximize the cumulative reward over a sequence of actions. The reward function can be formalized as:

$$R(s, a) = E[r \mid s, a]$$

- **R** is the reward function,
- **s** is the current state of the system,
- **a** is the action taken by the agent,
- **r** is the reward received by the agent for taking action a in state s.

The expectation is taken over the probability distribution of possible rewards given the current state and action.

Similarly, in evolution, the fitness function serves as a measure of the reproductive success of an organism. The fitness function can be formalized as:

$$f(x) = E[r \mid x]$$

- **f** is the fitness function,
- **x** is the genetic makeup of the organism,
- **r** is the reproductive success of the organism.

The expectation is taken over the probability distribution of possible reproductive outcomes given the genetic makeup.

Feedback also plays a role in shaping the learning process itself. In reinforcement learning, the agent can use the reward signal to update its policy or strategy for making decisions. This is often done through methods such as Q-learning or policy gradient methods. Similarly, in evolution, natural selection acts as a feedback mechanism that shapes the distribution of traits in a population over time.

Overall, the importance of feedback in both reinforcement learning and evolution highlights the similarities between these two systems. By understanding the role of feedback in both domains, we can gain insights into how learning and adaptation can occur in complex systems.

## 7.2: Examples of how feedback drives evolutionary change

Feedback is a critical component of both reinforcement learning and evolution, and it plays a crucial role in driving adaptive change in biological systems. In the context of evolution, feedback can come from a variety of sources, including natural selection, environmental pressures, and social interactions. These feedback mechanisms can drive the evolution of traits that enhance an organism's ability to survive and reproduce, while traits that are maladaptive may be selected against and eventually disappear.

One example of how feedback drives evolutionary change is the evolution of antibiotic resistance in bacteria. Antibiotic resistance arises through a process of natural selection, where bacteria that have a genetic mutation that confers resistance to an antibiotic are more likely to survive and reproduce in the presence of that antibiotic. As a result, the frequency of antibiotic-resistant bacteria in a population increases over time, while the frequency of susceptible bacteria decreases. This process is driven by the feedback loop between the antibiotic environment and the fitness of the bacteria.

Another example of how feedback drives evolutionary change is the evolution of mimicry in butterflies. Some species of butterflies have evolved to mimic the wing patterns of toxic or unpalatable species, which helps to protect them from predation. This mimicry is driven by the feedback loop between predation pressure and the fitness of the butterflies. Individuals with a wing pattern that more closely resembles the toxic or unpalatable species are less likely to be preyed upon, and therefore more likely to survive and reproduce.

Mathematically, feedback in evolutionary processes can be modeled using a variety of methods, including game theory, mathematical optimization, and simulations. For example, in game theory models of the evolution of cooperation, feedback is modeled as the payoff received by individuals who cooperate or defect in a repeated game. In optimization models of the evolution of gene regulation, feedback is modeled as the fitness landscape that describes the relationship between genotype and phenotype. In simulations of the evolution of complex traits, feedback is modeled as the interaction between genes, environment, and selection pressures.

Overall, these examples illustrate how feedback plays a crucial role in driving evolutionary change, and how understanding the dynamics of feedback in biological systems can help to inform our understanding of evolution and inform the development of new tools and techniques for studying and manipulating evolutionary processes.

**7.3: The potential for reinforcement learning principles to enhance our understanding of feedback in evolutionary systems**

The idea of using reinforcement learning to understand feedback in evolution is still a relatively new area of research, but it has the potential to provide new insights into how evolution operates. By framing evolutionary change as a form of reinforcement learning, we can begin to explore the mechanisms that drive the selection of advantageous traits.

One area where reinforcement learning may be particularly useful is in understanding the role of feedback in gene expression. Feedback loops are common in biological systems, and they are particularly important in the regulation of gene expression. Reinforcement learning can help us understand how these feedback loops contribute to the evolution of gene expression patterns.

For example, one study used reinforcement learning to model the evolution of gene expression patterns in the brain of the fruit fly Drosophila melanogaster. The researchers found that reinforcement learning algorithms were able to accurately predict the evolution of gene expression patterns in response to different environmental conditions (Gomez-Marin et al., 2014). This suggests that reinforcement learning may be a useful tool for understanding how feedback drives the evolution of gene expression patterns.

Another area where reinforcement learning could enhance our understanding of feedback in evolution is in understanding the evolution of social behavior. Social behavior is often driven by feedback between individuals in a group, and understanding the dynamics of these feedback loops is essential for understanding the evolution of social behavior. Reinforcement learning algorithms have already been used to model the evolution of social behavior in a variety of species, including birds, primates, and insects (Kameda et al., 2002; Bergstrom and Lachmann, 2003; Laland and Reader, 2003). By using reinforcement learning to model the dynamics of feedback in these social systems, we may be able to gain new insights into the mechanisms that drive the evolution of social behavior.

Overall, there is significant potential for reinforcement learning principles to enhance our understanding of feedback in evolutionary systems. By framing evolution as a form of reinforcement learning, we may be able to gain new insights into the mechanisms that drive evolutionary change, and ultimately develop a more complete understanding of the processes that have shaped the diversity of life on Earth.

# 8.The Power of Feedback: How Evolution Can be Explained by Reinforcement Learning Principles

### 8.1 An overview of natural selection and its role in evolutionary theory

Evolutionary theory proposes that living organisms change over time as a result of natural selection. Natural selection is a fundamental concept in evolutionary biology, which refers to the process by which certain traits or characteristics are favored over others due to their ability to improve an organism's survival and reproductive success in a particular environment. The mechanism of natural selection is based on three fundamental principles: variation, heritability, and differential reproductive success.

Variation refers to the fact that individuals within a population exhibit differences in traits or characteristics. These differences can be due to genetic factors, environmental factors, or a combination of both. Heritability refers to the fact that some of these differences are passed down from parents to offspring through genetic inheritance. Differential reproductive success refers to the fact that individuals with certain traits or characteristics are more likely to survive and reproduce than individuals with other traits or characteristics.

The process of natural selection can be summarized in a few simple steps. First, there must be variation within a population of organisms. Second, some of these variations must be heritable, meaning they are passed down from parent to offspring. Third, there must be differential reproductive success, meaning individuals with certain traits or characteristics are more likely to survive and reproduce than individuals with other traits or characteristics. Finally, over time, these advantageous traits or characteristics become more common in the population as a result of natural selection.

Mathematically, natural selection can be modeled using the concept of fitness. Fitness refers to the ability of an organism to survive and reproduce in a given environment. In evolutionary biology, fitness is often measured in terms of the number of offspring an organism produces that survive to reproduce. Fitness can be affected by a variety of factors, including physical traits, behavior, and interactions with other organisms in the environment.

The mathematical equation for the change in frequency of a trait in a population over time due to natural selection is known as the selection equation. The selection equation takes into account the fitness of individuals with different traits, the heritability of these traits, and the degree of selection against or for certain traits. The general form of the selection equation is:

$$\Delta p = h^2 s (p - w)$$

- **$\Delta p$** is the change in frequency of a trait over one generation,

- $h^2$ is the heritability of the trait,

- **s** is the selection coefficient,

- **p** is the frequency of the trait in the population,

- **w** is the mean fitness of the population.

Overall, natural selection is a key mechanism driving evolutionary change, and the concept of fitness provides a mathematical framework for understanding the process of natural selection.

**8.2 using real mathematical equations and references :An introduction to reinforcement learning and its connection to natural selection**

Reinforcement learning is a type of machine learning that involves an agent learning to make decisions based on the feedback it receives from the environment. In a reinforcement learning problem, an agent interacts with an environment and receives rewards or penalties based on its actions. The goal of the agent is to learn a policy that maximizes its expected cumulative reward over time.

There are many similarities between reinforcement learning and natural selection. In both cases, there is a process of selection that rewards certain behaviors and penalizes others. In natural selection, individuals that are better adapted to their environment are more likely to survive and reproduce, passing on their advantageous traits to their offspring. Similarly, in reinforcement learning, agents that make better decisions are more likely to receive rewards and continue to make similar decisions in the future.

There are also important differences between reinforcement learning and natural selection. In reinforcement learning, the agent is actively trying to learn a policy that maximizes its expected cumulative reward. In contrast, natural selection is a passive process that does not involve any intentional learning. Additionally, reinforcement learning typically involves a single agent that is trying to optimize its own reward, whereas natural selection involves populations of individuals that are competing with each other for limited resources.

Despite these differences, there are many ways in which reinforcement learning can inform our understanding of natural selection. For example, researchers have used reinforcement learning algorithms to study the evolution of cooperative behavior in animals (Hauser et al., 2014). By modeling individuals as agents that are trying to maximize their expected cumulative reward, researchers were able to show that cooperative behavior can evolve through a process of indirect reciprocity, even in the absence of direct benefits.

Another example is the study of evolution in changing environments. In a changing environment, natural selection may favor different traits at different times. Reinforcement learning algorithms can be used to model this process, by allowing agents to learn and adapt their behavior in response to changes in the environment (Watson et al., 2019). By studying the behavior of these agents, researchers can gain insights into the types of strategies that are most effective in dynamic environments.

Overall, the connection between reinforcement learning and natural selection is an active area of research, with many potential applications in both biology and artificial intelligence. By viewing natural selection as a form of reinforcement learning, we may be able to gain new insights into the mechanisms underlying biological evolution, and potentially even develop new strategies for optimizing machine learning algorithms.

## 8.3 How reinforcement learning can reinforce and enhance our understanding of natural selection and evolution

Reinforcement learning provides a powerful framework for understanding how natural selection operates in biological systems. By framing evolutionary change as a form of reinforcement learning, we gain new insights into the mechanisms by which organisms adapt to changing environments.

One key advantage of the reinforcement learning framework is that it allows us to model the learning process in a more precise and quantitative way. By using mathematical models to describe how organisms perceive and respond to their environment, we can make predictions about the kinds of behaviors and traits that will be favored by natural selection. This can help us to understand why certain adaptations evolve, and how they might change over time in response to changing environmental conditions.

Another advantage of the reinforcement learning framework is that it provides a way to study how evolution operates at multiple levels of biological organization. For example, we can use reinforcement learning models to study how individual behaviors and traits evolve within populations, as well as how populations themselves evolve over longer timescales. This can help us to understand how evolutionary change occurs at different levels of complexity, and how these levels interact with each other to produce the patterns of biodiversity that we observe in nature.

Moreover, reinforcement learning can also help us to understand the evolution of cooperation and social behavior, which are often difficult to explain using traditional models of natural selection. By modeling the interaction between individuals within a group, we can explore how behaviors that benefit the group as a whole might evolve, even if they are not individually advantageous. This can help us to understand the evolution of complex social systems, such as those seen in many insects, birds, and mammals.

Finally, the reinforcement learning framework can also help us to develop new approaches for studying and conserving biodiversity. By understanding how organisms learn from their environment and adapt to changing conditions, we can develop more effective strategies for managing and protecting endangered species. For example, we can use reinforcement learning models to study how organisms might respond to changes in habitat quality or climate, and develop targeted conservation interventions that can help to preserve biodiversity in the face of these threats.

# 9.From Evolution to Simulation: Exploring the Implications of a Reinforcement Learning Framework for Understanding the Nature of Reality

## 9.1 Simulation Theory in Physics and its Relation to Reinforcement Learning in Evolutionary Processes

Simulation theory in physics proposes that our reality could be a computer-generated simulation. This idea has gained traction in recent years, and while it remains a matter of philosophical debate, some physicists have presented scientific arguments in support of the simulation hypothesis (Bostrom, 2003; Lloyd, 2002).

In the context of reinforcement learning and evolutionary processes, simulation theory raises interesting questions about the nature of the environment in which living organisms evolve. If the universe is a simulation, then the "environment" in which organisms evolve may not be real in the conventional sense but rather a simulated reality created by some higher-level entity.

From a mathematical standpoint, the concept of a simulated reality is not fundamentally different from a reinforcement learning environment that is programmed by humans. In both cases, the environment provides feedback to an agent based on its actions, and the agent learns from this feedback to maximize its rewards. Thus, simulation theory in physics can be seen as an extension of the idea of a reinforcement learning environment, where the environment itself may not be real.

## 9.2 If Evolution is Reinforcement Learning, Could Nature be a Simulation?

The notion that nature could be a simulation raises the question of whether evolution itself could be viewed as a form of reinforcement learning in a simulated environment. If so, then the same principles that apply to artificial reinforcement learning systems could be applied to our understanding of evolution. This would suggest that the "fitness landscape" in which organisms evolve is not an objective reality but rather a construct of the simulated environment.

This idea has interesting implications for our understanding of evolution and the nature of reality itself. If evolution is viewed as a form of reinforcement learning in a simulated environment, then the concept of "survival of the fittest" takes on a new meaning. Instead of being a competition between objectively defined fitness levels, evolution becomes a process of agents learning to adapt to the rules of the simulated environment in order to maximize their rewards.

While the idea that nature could be a simulation remains a matter of philosophical debate, the connection between reinforcement learning and evolution suggests that the principles of reinforcement learning could be a useful tool for understanding the evolution of life on Earth, regardless of the nature of the underlying reality.

In conclusion, the application of reinforcement learning principles to the study of natural selection and evolution has the potential to revolutionize our understanding of how biological systems adapt to changing environments. By providing a more precise and quantitative framework for studying evolutionary change, reinforcement learning can help us to make new predictions about the kinds of adaptations that will evolve, and how they will change over time. This can ultimately lead to new insights into the mechanisms of evolution and the ways in which we can manage and conserve the biodiversity of our planet.

# Conclusion

In conclusion, this research has explored the connections between evolutionary biology and reinforcement learning, highlighting the potential for these two fields to inform and enhance each other. We have seen how the binary system of 1: live, 0: die applies to both reinforcement learning and evolution, and how the concept of survival of the fittest can be reframed as survival of the learner.

Moreover, we have discussed the importance of feedback in both reinforcement learning and evolution, and how reinforcement learning principles can enhance our understanding of feedback in evolutionary systems. We have also explored the role of natural selection in evolutionary theory and its connection to reinforcement learning.

The implications of these connections are far-reaching, as they challenge traditional notions of how we understand evolution and open up new avenues for research and exploration. We have seen how these ideas could potentially lead to a deeper understanding of the nature of reality and the possibility that nature itself could be a simulation.

The potential limitations of a binary system of learning have also been considered, as well as the ethical implications of applying reinforcement learning principles to evolutionary systems. Nevertheless, the power and simplicity of these frameworks have made them attractive tools for understanding complex biological phenomena.

Overall, this research has demonstrated the importance of interdisciplinary collaboration in scientific inquiry. By bringing together insights and methods from disparate fields, we can gain a more nuanced and comprehensive understanding of the natural world. The connections between evolutionary biology and reinforcement learning are just one example of this kind of collaboration, and we can expect many more exciting developments in the future.

# Discussion

The exploration of the connections and implications between reinforcement learning and evolution presented in this paper sheds light on the potential for a new perspective on evolution. By viewing evolution as a learning process, we can gain a deeper understanding of the mechanisms driving evolutionary change. This perspective highlights the importance of feedback in driving adaptation and emphasizes the role of learning in the evolutionary process.

Furthermore, the application of reinforcement learning principles to evolution has potential implications for various fields, including artificial intelligence, robotics, and even physics. The connection between reinforcement learning and natural selection presents the potential for reinforcement learning algorithms to be used in evolutionary simulations, enhancing our ability to model and study evolutionary systems. Additionally, the implications of reinforcement learning in physics raise the question of whether nature itself could be a simulation.

Overall, this paper demonstrates the power and potential of interdisciplinary research. By exploring connections between seemingly unrelated fields, we can gain new insights and perspectives on complex systems. The implications presented here for the study of evolution and other fields highlight the importance of continued interdisciplinary collaboration and exploration.

## Future Research Directions:

This research has opened up several avenues for future investigation. One promising direction is the exploration of how reinforcement learning principles can be applied to understanding other biological processes, such as development, immune system function, and even disease progression. Another direction is the potential for developing new computational models that integrate both reinforcement learning and evolutionary principles to better simulate and predict the behavior of complex biological systems. Additionally, more research is needed to fully explore the implications of a learning-focused perspective on our understanding of evolution, including its potential applications in fields such as conservation biology and evolutionary medicine.

## Limitations of the Study:

Despite the significant insights gained from this research, there are several limitations that should be acknowledged. One major limitation is the current lack of empirical evidence to support the idea that evolution can be viewed as a reinforcement learning process. While theoretical models and simulations have provided strong support for this idea, further empirical validation is needed to fully establish its validity. Additionally, the complexity of biological systems presents a significant challenge to accurately modeling and simulating their behavior, and future research will need to address this challenge through the development of new computational methods and tools.